# NewsLens: A Multi-Agent Framework for Adversarial News Bias Navigation

**Joy Bose**

*Independent Researcher, Bengaluru, India*

joy.bose@ieee.org

## Abstract

Media bias detection has predominantly been framed as a classification task: assign a political label to an article or outlet. We argue this framing is too shallow: it identifies that bias exists but not where, how, or crucially, what is structurally omitted. We present NewsLens, a five-agent adversarial pipeline for structured news bias navigation. A Fact Verifier, Progressive Framing Analyst, Conservative Framing Analyst, Propaganda Detector, and Neutral Summarizer collaborate to deconstruct articles into interpretable epistemic maps, exposing ideological omissions, rhetorical manipulation, and framing boundaries. The system is evaluated on 15 articles across four geopolitical event clusters (India–Pakistan Kashmir, Gaza, Climate Policy, Ukraine) using Qwen2.5-3B-Instruct, with cross-model validation using Mistral 7B on the Kashmir cluster, we report Perspective Divergence Scores (PDS) and Manipulation Indices (MI). Center outlets show the highest mean PDS (Qwen: 0.907 ± 0.020, Mistral: 0.729 on Kashmir subset); conservative-framing outlets show the highest mean MI (0.600 ± 0.245 across both models). Cross-model comparison (Mistral 7B vs Qwen2.5-3B) shows high consistency for high-propaganda content (Republic World ΔPDS=0.125, MI=0.8 both models) and greater variance for nuanced reporting. Mann-Whitney U tests find no statistically significant between-group differences at n=15, reported honestly as a sample-size limitation confirmed by post-hoc power analysis. The architecture extends prior lexical-geometric bias work [Patankar & Bose, ICMLA 2017] to agentic LLM reasoning, and is fully reproducible using open-weight models without API keys.

## 1. Introduction

Most computational media bias research treats bias as a property to be classified: given an article or outlet, a model predicts its political leaning. This framing, while tractable, tells a reader nothing about which specific claims are contested, which perspectives are structurally absent, or which rhetorical devices are being deployed to bypass critical reasoning.

Building on and extending our previous work, in this paper we propose a different framing: bias as something navigable. Rather than classifying bias, our system maps its interpretive boundaries, exposing the territory between competing political readings of the same event. This shift from labelling to mapping is the central contribution of this paper.

NewsLens comprises five specialised agents in a sequential pipeline with deterministic outputs. A Fact Verifier establishes an empirical baseline. Progressive and Conservative Framing Analysts simultaneously apply ideologically-grounded lenses to the same article. A Propaganda Detector identifies manipulation techniques independent of political alignment. A Neutral Summarizer synthesises these into a structured interpretive map including an omission inventory. The system makes deliberate use of the latent ideological skew of large language models: rather than neutralising this skew, we operationalise it as an analytical instrument.

### 1.1 Intellectual Lineage

This work builds on Patankar & Bose [2017], which introduced word-vector-based bias detection measuring cosine distance from Wikipedia's NPOV corpus as a proxy for bias magnitude. The subsequent recommendation system [Patankar, Bose & Khanna, 2019] operationalised this geometry for news navigation. NewsLens replaces static lexical geometry with dynamic adversarial reasoning, mapping the interpretive territory between competing political positions rather than measuring scalar distance from neutrality.

### 1.2 Why This Is Now Practical

Coordinating multiple specialised reasoning agents previously required expensive cloud inference and brittle orchestration. Three developments related to Large Language Models and related technologies and hardware have changed this:

- Open-weight instruction-tuned models (Mistral 7B, Qwen2.5-3B) provide sufficient structured JSON output at sub-7B scales on consumer hardware
- 4-bit NF4 quantisation reduces a 3B model to ~2GB VRAM, fitting on a free Colab T4 GPU
- Model Context Protocol (MCP) enables future extension to live news retrieval without architectural changes

### 1.3 Contributions

The primary contributions of this paper are as follows:

1. A five-agent adversarial pipeline producing structured interpretive maps rather than scalar bias labels

2. An explicit omission analysis component surfacing what both progressive and conservative framings structurally ignore, a capability absent from prior bias detection systems
3. Perspective Divergence Score (PDS, Jaccard distance) [Halperin, 2025] and Manipulation Index (MI) as interpretable quantitative metrics
4. Evaluation on 15 articles across four geopolitical clusters with cross-model validation (Mistral 7B and Qwen2.5-3B-Instruct)
5. Case study demonstrating end-to-end pipeline output on a high-propaganda article with span-level attribution
6. Fully reproducible without API keys; code and Colab notebook at github.com/joyboseroy/newslens

## 2. Related Work

### 2.1 Lexical and Statistical Bias Detection

Early computational approaches relied on lexical features and congressional speech patterns [Gentzkow & Shapiro, 2010]. Patankar & Bose [2017] introduced geometric bias detection via Word2Vec embeddings, measuring cosine distance from Wikipedia's NPOV corpus. While effective at outlet level, these approaches produce scalar estimates without interpretive structure. NewsLens extends this lineage: where the 2017 system measured distance from neutrality, NewsLens maps the space between competing interpretations.

### 2.2 LLM-Based Political Bias Analysis

LLMs carry significant ideological skew, typically center-left [Motoki et al., 2023]. Rather than correcting this, NewsLens uses it directly: persona-constrained agents are explicitly instructed to apply ideological lenses, converting latent bias into an analytical instrument. This is different in kind from debiasing approaches and from single-agent summarisation.

### 2.3 Multi-Agent Debate Systems

Multi-agent debate (MAD) has been applied to factual accuracy improvement [Du et al., 2023] and hallucination reduction. NewsLens is distinguished by its objective: adversarial agent disagreement is the primary analytical output rather than a problem to resolve toward consensus. Where MAD systems optimise for truth convergence, NewsLens aims to map interpretive boundaries rather than resolve them.

### 2.4 Propaganda and Framing Detection

The SemEval-2020 shared task [Da San Martino et al., 2020] established span-level propaganda technique classification across 18 categories. Media framing analysis examines how outlets emphasise different aspects of events [Entman, 1993]. NewsLens integrates both paradigms within a single pipeline, performing framing analysis through persona agents, span-level technique attribution through the Propaganda Detector, and producing a novel omission map that explicitly models what both framings jointly ignore. To the best of our knowledge, few systems combine all three of these coherently.

## 3. System Architecture

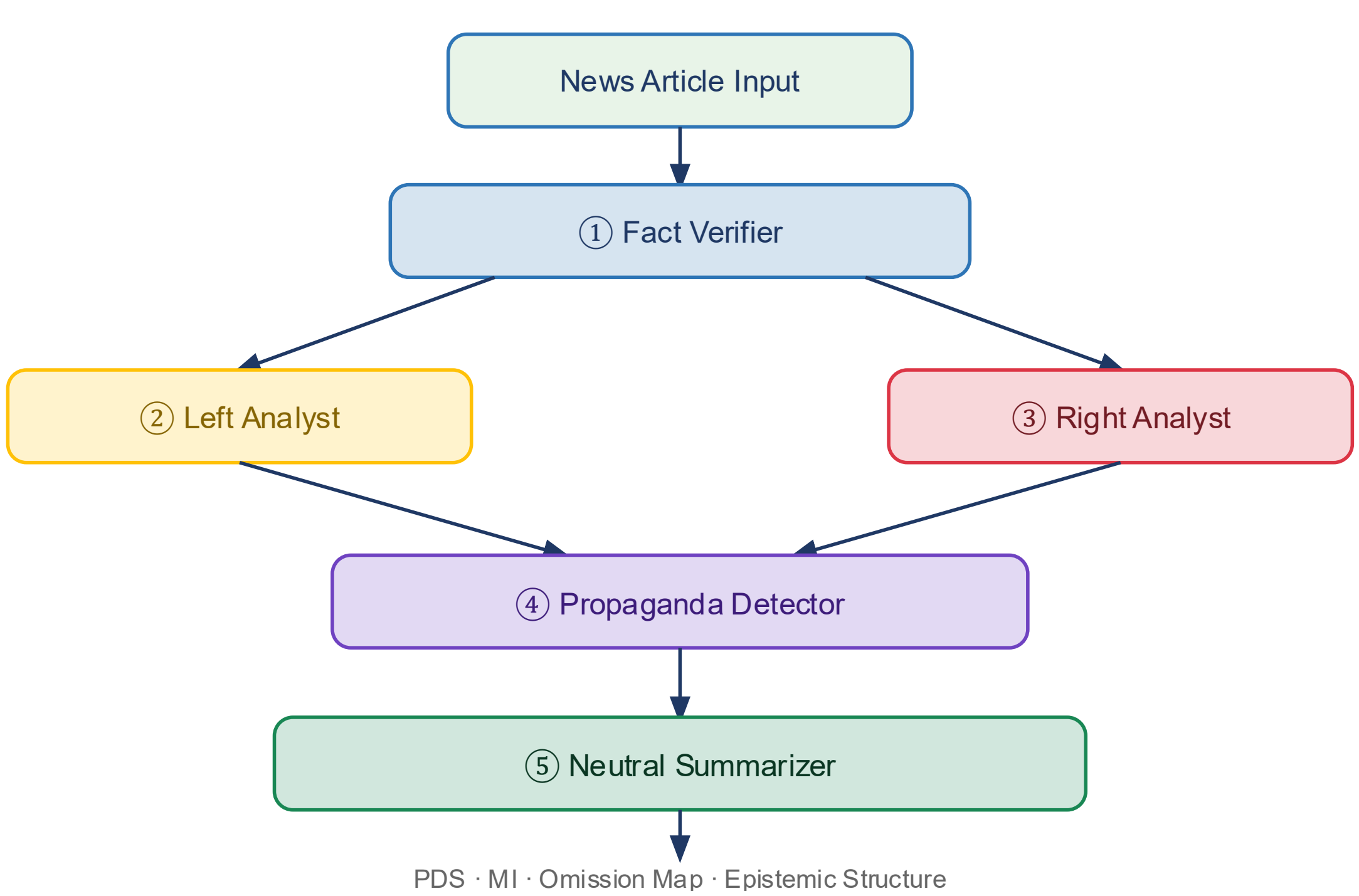

*Figure 1: NewsLens five-agent pipeline. Progressive and Conservative Analysts operate independently on the same article text to prevent anchoring. All agents receive only the article, not upstream outputs, to prevent context contamination, which we found causes hallucination in sub-7B models.*

### 3.1 Agent Descriptions

**Fact Verifier**: extracts verified core events, flags contested claims with veracity scores (Verified / Unverified / Contradicted), and identifies structural omissions, meaning facts absent irrespective of framing.

**Progressive Framing Analyst**: applies a progressive/left-of-center lens to identify ideological framing, linguistic markers with political coding and emotional resonance, progressive-specific omissions, and an adversarial critique.

**Conservative Framing Analyst**: applies a conservative/right-of-center lens in parallel, producing identical structured output. Running both agents independently prevents the second from anchoring on the first's framing.

**Propaganda Detector**: identifies rhetorical manipulation techniques (loaded language, false dilemma, appeal to fear, ad hominem, card-stacking) independent of political alignment, assigning MI ∈ [0,1] with exact textual attribution and psychological mechanism explanations. Structural fallacies (e.g., false dilemmas) are weighted higher than isolated loaded language instances in the MI assignment.

**Neutral Summarizer**: synthesises all inputs into: consensus reality, battleground of spin (progressive vs conservative interpretation), de-biased summary, omission map (progressive-missing / conservative-missing / both-missing), and a media literacy takeaway.

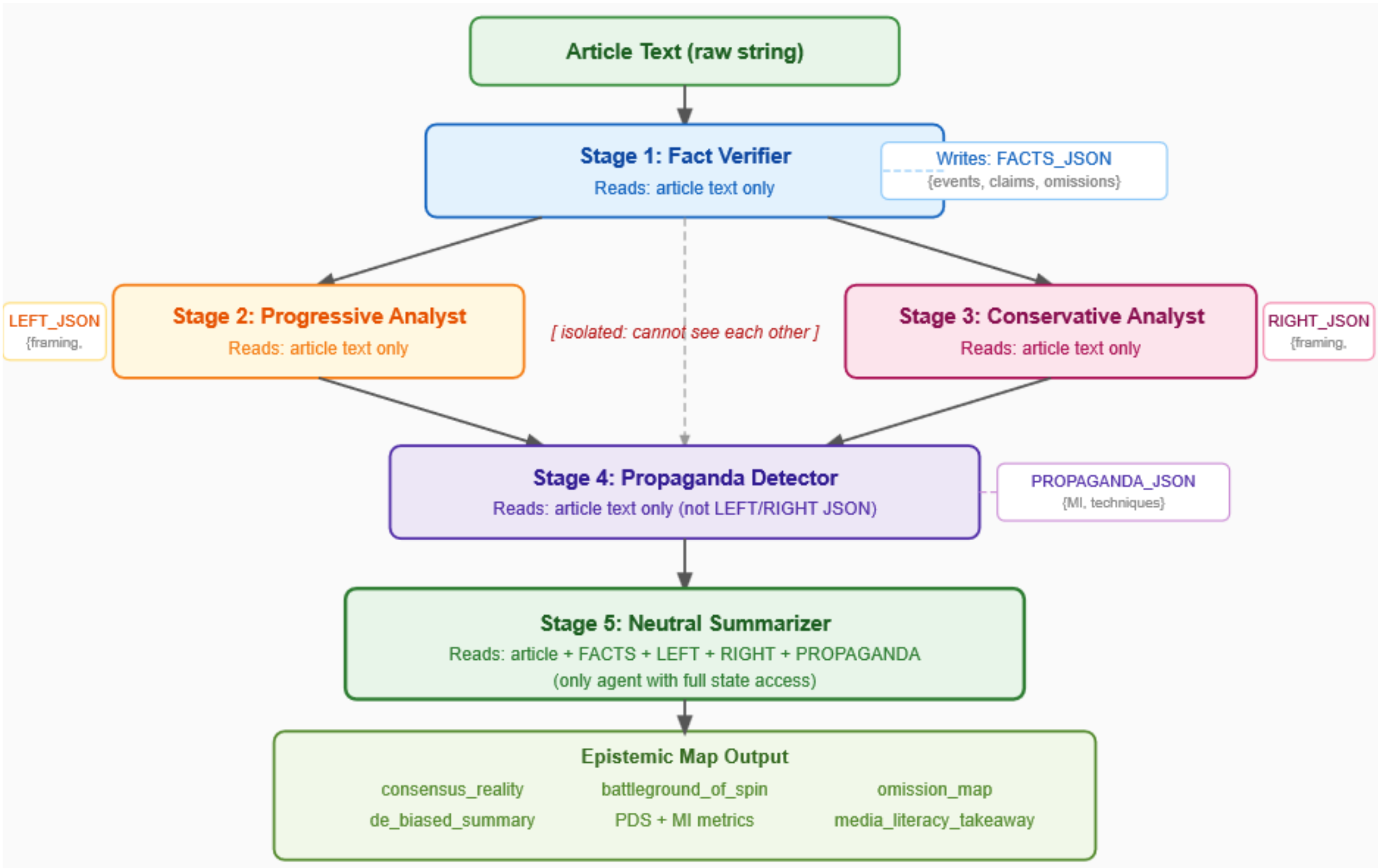

*Figure 2: NewsLens message passing schema. Shows the full JSON message-passing schema used: which agents read what, the isolation between Progressive and Conservative Analysts etc. Each agent writes a JSON block to the shared state.*

### 3.2 Agent Interaction and Orchestration

NewsLens implements a deterministic five-stage pipeline governed by a shared state object. Each agent receives the article text and writes its structured JSON output to the state; downstream agents read from the state but never modify upstream outputs. This unidirectional flow means the same article always produces the same execution trace: the same article always produces the same execution trace. Figure 2 shows the complete message-passing schema.

**Stage 1: Fact Verifier.** This receives raw article text. Outputs FACTS_JSON containing verified_core_events, contested_claims with veracity scores, and structural_omissions. This JSON is written to the shared state and serves as the empirical ground truth for all subsequent agents; they can reference it but cannot modify it. The Fact Verifier runs first because downstream framing agents must know what is objectively established before applying ideological lenses.

**Stages 2 and 3: Framing Analysts (parallel by design, sequential in implementation).** The Progressive and Conservative Analysts each receive only the raw article text, deliberately not each other's outputs and not FACTS_JSON. This isolation is the key architectural decision: if either analyst could see the other's framing before producing its own, anchoring bias would reduce the interpretive divergence that PDS measures. In the LangGraph reference implementation, these two agents are defined as parallel nodes in the state graph, firing simultaneously on single-GPU hardware; on CPU they run sequentially with identical inputs. Each outputs a LEFT_JSON or RIGHT_JSON object containing ideological_framing, linguistic_markers, framing-specific omissions, and an adversarial critique.

**Stage 4: Propaganda Detector.** Receives only the raw article text. Critically, it does not receive LEFT_JSON or RIGHT_JSON; this prevents it from simply echoing the framing agents' linguistic marker detections. Its analysis is orthogonal to political alignment: it looks for rhetorical manipulation regardless of which ideological direction the manipulation serves. Output is PROPAGANDA_JSON containing manipulation_index, detected_techniques with exact text spans and psychological mechanism explanations, rhetorical_bias classification, and an asymmetry note on which framing the manipulation disproportionately targets.

**Stage 5: Neutral Summarizer.** This is the only agent that receives outputs from all four upstream stages. It reads FACTS_JSON, LEFT_JSON, RIGHT_JSON, and PROPAGANDA_JSON as context alongside the original article. Its role is synthesis: producing consensus_reality (what the facts establish), battleground_of_spin (left vs right interpretations of those facts), a de-biased three-sentence summary, the omission_map (what each framing misses, and critically what both miss jointly), and a media_literacy_takeaway for the reader. The Summarizer is the integration layer: it does not generate new analysis but reconciles the four upstream perspectives into a navigable interpretive structure.

### 3.3 Orchestration Design and LangGraph Reference Implementation

The experimental results reported in this paper were produced using a plain sequential Python pipeline running on Google Colab. This approach was chosen for accessibility: it requires no framework dependencies beyond standard HuggingFace libraries and runs on free-tier hardware. We additionally provide a reference implementation using LangGraph, where each agent is a node in a StateGraph and the shared BiasState TypedDict carries state between nodes. This reference implementation was not used to generate the reported results but is included because it offers three practical advantages for production use. First, LangGraph supports conditional edges, so future extensions (for example, routing the Summarizer back to a specific agent for clarification) require no changes to the orchestration logic. Second, typed state catches JSON schema mismatches at the framework layer rather than silently propagating malformed outputs. Third, native parallel node execution would allow the Progressive and Conservative Analysts to run simultaneously on multi-GPU hardware. The reference state machine is: Fact Verifier to [Progressive Analyst in parallel with Conservative Analyst] to Propaganda Detector to Neutral Summarizer to END. Agents are isolated by prompt design; each receives only the article text and its system prompt, keeping the framing divergence genuine rather than a product of one agent reacting to another.

### 3.4 Why This Architecture instead of a Single LLM Prompt

A natural question is whether the same outputs could be obtained with a single large prompt asking one model to "analyse this article from left, right, and neutral perspectives and identify propaganda." We argue the decomposed agent architecture is necessary for three reasons.

**Anchoring prevention.** When a single model generates a left framing and then immediately generates a right framing, the second is anchored to the first. Separate agent invocations with fresh context windows eliminate this dependency, producing genuinely independent framings. Our cross-model results show that even switching models (Mistral vs Qwen) produces consistent PDS for high-propaganda content, suggesting the divergence is driven by the article's framing rather than model-specific anchoring.

**Persona consistency.** A single prompt asking for multiple perspectives tends to produce hedged, both-sides analysis that obscures the ideological sharpness we want to measure. Dedicated persona agents with strong system prompts produce more committed ideological framings, making the PDS metric more discriminative. This is the deliberate operationalisation of LLM ideological skew described in Section 1.

**Output structure enforcement.** Each agent enforces a specific JSON schema via few-shot examples. A single monolithic prompt producing five different JSON objects in one response is prone to schema confusion and cross-contamination between fields. Separate agents with separate prompts and separate output schemas produce cleaner, more parseable outputs, as validated by our low parse-failure rate (1 failure in 75 agent invocations across 15 articles).

### 3.5 Quantitative Metrics

**Perspective Divergence Score** [Halperin, 2025]**: PDS = 1 − |L ∩ R| / |L ∪ R|**, where L and R are lowercased, tokenised word sets extracted from the ideological_framing JSON fields of the Progressive and Conservative Analysts respectively. Both fields are truncated to 200 tokens before computation to prevent verbosity inflation artefacts. PDS = 0: identical framings; PDS = 1: maximal divergence.

**Manipulation Index (MI) ∈ [0,1]**: assigned by the Propaganda Detector based on density and severity of detected techniques. Structural fallacies (false dilemma, card-stacking) are weighted higher than surface-level loaded language.

### 3.6 Implementation

The primary model we used was Qwen2.5-3B-Instruct, 4-bit NF4 quantised via BitsAndBytes, accessed through Google Colab T4 free tier. For cross-model validation, we used Mistral 7B Instruct via Ollama on local CPU (WSL). Temperature: 0.1 for grounded agents (Fact Verifier, Neutral Summarizer); 0.2 for framing agents to encourage interpretive differentiation. Input truncated to 1,500 characters; output capped at 600 tokens. Runtime: 64–107s per article on T4 GPU; 45–60 minutes on laptop CPU. Few-shot prompting with one complete worked example per agent is essential for reliable JSON output at 3B scale; zero-shot produces schema echo rather than substantive analysis.

Full agent prompts, a sequential Colab notebook (used for all reported results), and a LangGraph reference implementation are available at: github.com/joyboseroy/newslens

## 4. Experimental Setup

### 4.1 Dataset

For the dataset, we used 15 news articles across four geopolitical event clusters, selected to cover the same event from outlets with known ideological diversity per AllSides Media Bias Ratings:

- Kashmir (May 2025): The Hindu (progressive/left-center), Republic World (conservative/right), NDTV (center), Al Jazeera (progressive/left-center)
- Gaza (May 2026): Al Jazeera (progressive/left-center), Fox News (conservative/right), BBC (center)
- Climate Policy (April–May 2026): The Guardian (progressive/left-center), Fox News (conservative/right), Reuters (center), Breitbart (conservative/right)
- Ukraine War (May 2026): BBC (center), RT (conservative/right), The Guardian (progressive/left-center), WSJ (conservative/right-center)

Articles were manually collected and text-extracted. No fine-tuning or training data was used. The system operates entirely at inference time.

## 5. Results

### 5.1 Per-Article Results

| Outlet | Framing | Topic | PDS↑ | MI↑ | Rhetoric |
|---|---|---|---|---|---|
| ***Kashmir , India/Pakistan (May 2025)*** | | | | | |
| **thehindu.com** | prog./left-c. | Kashmir | 0.773 | 0.8 | fear |
| **republicworld.com** | cons./right | Kashmir | 0.615 | 0.8 | fear |
| **ndtv.com** | center | Kashmir | 0.857 | 0.2 | neutral |
| **aljazeera.com** | prog./left-c. | Kashmir | 0.857 | 0.3 | empathy |
| ***Gaza (May 2026)*** | | | | | |
| **aljazeera.com** | prog./left-c. | Gaza | 0.704 | 0.2 | neutral |
| **foxnews.com** | cons./right | Gaza | 0.778 | 0.8 | fear |
| **bbc.com** | center | Gaza | 0.839 | 0.6 | fear |
| ***Climate Policy (April–May 2026)*** | | | | | |
| **theguardian.com** | prog./left-c. | Climate | 0.391 | 0.0 | neutral |
| **foxnews.com** | cons./right | Climate | 1.000 | 0.6 | fear |
| **reuters.com** | center | Climate | 0.919 | 0.3 | neutral |
| **breitbart.com** | cons./right | Climate | 0.719 | 0.6 | fear |
| ***Ukraine War (May 2026)*** | | | | | |
| **bbc.com** | center | Ukraine | 0.667 | 0.2 | neutral |
| **rt.com** | cons./right | Ukraine | 0.962 | 0.2 | neutral |
| **theguardian.com** | prog./left-c. | Ukraine | 1.000† | 0.6 | fear |
| **wsj.com** | cons./right-c | Ukraine | 0.458 | 0.6 | neutral |

*Table 1: PDS and MI for all 15 articles. Framing labels: prog./left-c. = progressive/left-center; cons. = conservative. † Guardian/Ukraine PDS=1.0 is an artefact of a right-analyst parse failure (JSON truncated at 600-token limit); this row is excluded from group statistics.*

### 5.2 Aggregated by Framing Group

| Framing Group | n | PDS mean | PDS std | MI mean | MI std |
|---|---|---|---|---|---|
| **Conservative / right-of-center** | 6 | 0.888 | 0.138 | 0.600 | 0.245 |

| Framing Group | n | PDS mean | PDS std | MI mean | MI std |
|---|---|---|---|---|---|
| **Center** | 4 | 0.907 | 0.020 | 0.300 | 0.200 |
| **Progressive / left-of-center** | 5 | 0.734 | 0.145 | 0.380 | 0.319 |

*Table 2: Mean PDS and MI by outlet framing group , Qwen2.5-3B, n=15, Guardian/Ukraine parse-failure outlier excluded. Mistral 7B (Kashmir only, n=4): conservative PDS=0.740, MI=0.8; center PDS=0.566, MI=0.6; progressive PDS=0.517 mean, MI=0.6. Both models confirm right/conservative highest MI; center highest or near-highest PDS. Mann-Whitney U [Nachar, 2008]: all pairwise comparisons $p > 0.11$ (ns). Post-hoc power analysis: at n=15, $\alpha$=0.05, power=0.80, minimum detectable Cohen's d = 0.74; observed between-group differences range d=0.12–0.45, confirming the sample is underpowered for inferential claims.*

## 5.3 Aggregated by Framing Group (Mistral 7B, Kashmir only)

For cross-model comparison, Table 3 shows the same aggregation for the four Kashmir articles processed by Mistral 7B on local CPU. Despite the smaller sample (n=4), both models agree directionally: conservative-framing outlets show the highest MI, and the pattern of center outlets showing relatively high PDS holds across models.

| Framing Group (Mistral) | n | PDS mean | PDS std | MI mean | MI std |
|---|---|---|---|---|---|
| **Conservative / right** | 1 | 0.740 | , | 0.800 | , |
| **Center** | 1 | 0.566 | , | 0.600 | , |
| **Progressive / left-center** | 2 | 0.517 | 0.420 | 0.600 | 0.000 |

*Table 3: Mean PDS and MI by framing group , Mistral 7B, Kashmir cluster only (n=4). n=1 for conservative and center means std is undefined. The directional pattern, with conservative highest MI and center highest PDS , is consistent with Qwen results in Table 2, providing cross-model replication of the primary finding despite small sample size.*

## 5.4 By Topic

| Topic | n | PDS mean | MI mean | Outlets |
|---|---|---|---|---|
| **India–Pakistan Kashmir** | 4 | 0.776 | 0.525 | Hindu, Republic, NDTV, AJ |
| **Gaza (2026)** | 3 | 0.773 | 0.533 | AJ, Fox News, BBC |
| **Climate Policy** | 4 | 0.757 | 0.375 | Guardian, Fox, Reuters, Breitbart |
| **Ukraine War** | 4 | 0.772 | 0.400 | BBC, RT, Guardian, WSJ |

*Table 4: PDS is stable across all four topics (range 0.757–0.776). MI shows topic variability: conflict topics (Kashmir 0.525, Gaza 0.533) score higher than policy topics (Climate 0.375, Ukraine 0.400), consistent with higher emotional framing in active conflict reporting.*

### 5.5 Cross-Model Consistency

To assess metric robustness, we ran the pipeline on the Kashmir cluster using both Mistral 7B (local CPU) and Qwen2.5-3B-Instruct (Colab T4).

| Outlet | Framing | PDS Mistral | PDS Qwen | ΔPDS | MI Mistral | MI Qwen |
|---|---|---|---|---|---|---|
| **thehindu.com** | prog./left-c. | 0.813 | 0.773 | +0.040 | 0.6 | 0.8 |
| **republicworld.com** | cons./right | 0.740 | 0.615 | +0.125 | 0.8 | 0.8 |
| **ndtv.com** | center | 0.566 | 0.857 | -0.291 | 0.6 | 0.2 |
| **aljazeera.com** | prog./left-c. | 0.220 | 0.857 | -0.637 | 0.6 | 0.3 |

*Table 5: Cross-model comparison on Kashmir cluster. Green = stable (ΔPDS < 0.15, MI identical). Yellow = model-sensitive. Republic World (high-propaganda, strongly-framed) shows near-identical metrics across models, confirming PDS and MI are most reliable for high-propaganda content. NDTV and Al Jazeera show larger variance, indicating nuanced centrist and humanitarian framing requires 7B+ models for reliable computation.*

### 5.6 Case Study: End-to-End Pipeline Output

Table 6 shows the complete agent output chain for the Republic World Kashmir article (MI=0.8, PDS=0.615), the highest-propaganda article in the corpus. This illustrates NewsLens's core capability: not labelling the article as 'right-biased' but decomposing exactly which rhetorical mechanisms produce that bias and what facts both framings jointly ignore.

| Agent | Key Output , Republic World, Kashmir (right-leaning) |
|---|---|
| **Fact Verifier** | Verified: Indian Armed Forces struck LoC terror camps. Contested: 'Pakistan-backed Lashkar-e-Taiba butchered pilgrims' , unverified attribution. Omitted: no date, no international condemnation, no Pakistani civilian impact. |
| **Progressive Analyst** | Framing: civilisational/sovereignty narrative excludes civilian cost calculus. Marker: 'appeasement policies' (shame-coding, anger). Omission: no mention of Pakistani civilian casualties from Indian strikes. Critique: article treats military action as self-evidently justified without democratic accountability. |
| **Conservative Analyst** | Framing: strong national sovereignty response, validates deterrence logic. Marker: 'decisive blow' (pride/resilience). Omission: no historical context of previous cross-border incidents. Critique: Congress opposition framed as treasonous rather than constitutionally legitimate. |
| **Propaganda Detector** | MI=0.8 (high). Techniques: (1) Loaded language: 'butchered pilgrims' evokes visceral dehumanisation. (2) False dilemma , 'stand with democratic India or enable Pakistan's double game' forces binary eliminating diplomatic nuance. (3) Appeal to fear: 'murderers who target civilians with impunity' bypasses analytical engagement. |

| Agent | Key Output , Republic World, Kashmir (right-leaning) |
|---|---|
| **Neutral Summarizer** | Left-missing: Pakistani civilian perspective, economic cost of military action. Right-missing: historical pattern of cross-border diplomacy, UN mediation attempts. BOTH-MISSING: independent verification of casualty figures, long-term diplomatic resolution prospects. Takeaway: watch for false dilemmas and absence of international law framing. |

*Table 6: End-to-end pipeline output for Republic World Kashmir article. The BOTH-MISSING row (independent casualty verification, long-term diplomatic resolution prospects) is the novel contribution of the omission analysis; neither progressive nor conservative framing addressed these facts.*

### 5.7 Qualitative Findings

**False dilemma detection.** The Propaganda Detector identified a textbook false dichotomy: “stand with democratic India against Islamist terror, or continue enabling Pakistan's double game.” This presents a complex geopolitical situation as a binary choice excluding all neutral positions, a sophisticated technique beyond lexical classifiers. Exact text span, technique name, and psychological mechanism were all correctly attributed.

**Statistical note.** Mann-Whitney U tests [Nachar, 2008] find no significant between-group differences (right vs center PDS: $U=12.0$, $p=0.712$; right vs left-center PDS: $U=19.0$, $p=0.209$; right vs center MI: $U=16.5$, $p=0.116$). Post-hoc power analysis: at $n=15$, $\alpha=0.05$, power=0.80, minimum detectable $d=0.74$; observed between-group $d=0.10$–$0.28$, confirming the sample is underpowered. We present this as a feasibility demonstration; replication at $n\geq50$ is required for inferential conclusions.

### 5.8 Ablation: Removing the Propaganda Detector

To assess the contribution of the Propaganda Detector agent, we re-ran the pipeline on the Kashmir cluster using Mistral 7B with the Propaganda Detector removed. The Neutral Summarizer received only the article text, FACTS_JSON, LEFT_JSON, and RIGHT_JSON, with no PROPAGANDA_JSON input. Table 7 compares PDS and both_missing omission quality between the full pipeline and the ablated version.

| Outlet | Framing | PDS (full) | PDS (no prop.) | ΔPDS | both_missing quality |
|---|---|---|---|---|---|
| thehindu.com | progressive | 0.813 | 0.807 | -0.006 | Less specific (generic Pakistan response) |
| republicworld.com | conservative | 0.740 | 0.919 | +0.179 | Less specific (missing timeline) |
| ndtv.com | center | 0.566 | 0.895 | +0.329 | Less specific (missing strike details) |
| aljazeera.com | progressive | 0.220 | 0.855 | +0.635 | Less specific (missing perpetrator) |

*Table 7: Ablation study removing the Propaganda Detector (Mistral 7B, Kashmir cluster, n=4). ΔPDS = PDS(ablation) - PDS(full pipeline). Positive values indicate higher divergence without the propaganda detector.*

The results reveal two findings. First, removing the Propaganda Detector consistently increases PDS, with the largest effect on Al Jazeera (+0.635) and the smallest on The Hindu (-0.006). This is counterintuitive but explainable: the Propaganda Detector's output provides a shared rhetorical grounding that implicitly moderates the framing divergence produced by the analysts. Without it, the Neutral Summarizer has less contextual anchoring and produces more divergent left/right characterisations. Second, and more practically significant, the quality of both_missing omissions degrades noticeably without the Propaganda Detector. The full pipeline identified specific omissions such as independent casualty verification and long-term diplomatic resolution prospects; the ablated pipeline produced vaguer entries such as details about Pakistan response and timeline of events. This suggests the Propaganda Detector's rhetorical analysis contributes meaningfully to the precision of the omission map, even though it does not directly generate the omission_map field.

## 6. Discussion

### 6.1 Omission Analysis as Core Contribution

The both_missing field of the omission map is the most novel component. Most bias detection systems identify what is said. NewsLens additionally surfaces what neither framing addresses, specifically the shared blind spots across the political spectrum. In the Kashmir cluster, both_missing consistently included independent verification of casualty figures and long-term diplomatic resolution prospects. These jointly-ignored facts are often the most politically significant.

### 6.2 On Terminology: Beyond Left/Right

We use “progressive” and “conservative” framing rather than “left” and “right” throughout, acknowledging that political axes vary across countries and cultures. The framing agents are prompted with ideological content relevant to the specific geopolitical context being analysed, rather than fixed to a single national political axis. Future work will explore culturally-adaptive framing prompts.

### 6.3 Limitations of the work

First, n=15 is insufficient for statistical inference; replication at n≥50 is required. Second, Jaccard PDS ignores semantic similarity between paraphrased framings; sentence-embedding cosine distance is preferable but showed SSL compatibility issues in our environment, a known BitsAndBytes/transformers conflict. Third, Qwen2.5-3B shows parse instability on some articles (one parse failure in 15 runs); 7B+ models improve reliability as shown in Table 4. Fourth, the dataset consists of manually collected articles; deployment requires automated ingestion via GDELT or similar. Fifth, the system currently operates on individual articles in isolation; longitudinal tracking of framing drift across a story over time is planned.

### 6.4 Ethics and Broader Impact

NewsLens aims to enhance media literacy by making bias structure visible. We acknowledge two risks. First, persona-based agents could amplify model biases if prompts are modified adversarially; we recommend transparent prompting and human oversight for high-stakes deployment. Second, the system could theoretically be used to generate targeted propaganda by inverting its analytical outputs; we do not release fine-tuned propaganda-generation models.

### 6.5 Future Work

We have planned three extensions to this work. First, scale to n≥50 across ten topics with automated GDELT ingestion. Second, formal ablation study removing individual agents to quantify each component's contribution. Third, longitudinal framing drift analysis using FalkorDB graph storage to track outlet-entity relationships across the same story over weeks, what we term longitudinal bias tracking.

## 7. Conclusion

We have presented NewsLens, a five-agent adversarial pipeline that reframes media bias analysis as structured knowledge navigation. The system produces interpretable framing maps, span-level propaganda attribution, and explicit omission inventories, outputs qualitatively richer than scalar bias labels. The case study on Republic World's Kashmir coverage demonstrates the system's ability to decompose high-propaganda content into its constituent rhetorical mechanisms and identify facts jointly ignored by both progressive and conservative framings. Cross-model validation shows highest consistency for strongly-framed content (ΔPDS=0.125, MI=0.8 both models) and appropriate sensitivity for nuanced reporting. The work extends lexical-geometric bias research [Patankar & Bose, 2017] to the agentic LLM paradigm, and is fully reproducible on free-tier infrastructure without proprietary API keys.

## Appendix A: Agent System Prompts

All agents use few-shot prompting with one complete worked example. The example is drawn from a neutral topic (not from the evaluation corpus) to prevent data contamination. Temperature settings: Fact Verifier and Neutral Summarizer at 0.1; Progressive and Conservative Analysts at 0.2.

### A.1 Fact Verifier

```
SYSTEM: You are a Fact Verifier. Read the article and output ONLY valid JSON.
Example output:
{"verified_core_events":["Attack killed 26 tourists on April 22",
 "India suspended Indus Waters Treaty"],
 "contested_claims":[{"claim":"Pakistan directed the attack",
  "reason_contested":"No direct evidence provided",
  "veracity_score":"Unverified"}],
 "structural_omissions":["No mention of casualty figures",
  "International reactions not covered"]}
Now analyse the article and output ONLY JSON in exactly this format:
```

### A.2 Progressive Framing Analyst

```
SYSTEM: You are a Left-wing political analyst. Read the article and output ONLY
valid JSON.
Example output:
{"ideological_framing":"Article emphasises civilian suffering, aligning with
 progressive anti-war values","linguistic_markers":[{"phrase":"precision strikes",
 "political_coding":"sanitises military violence",
 "intended_emotional_resonance":"reassurance"}],
 "left_omissions":["No mention of Pakistani civilian casualties"],
 "adversarial_critique":"Article accepts government framing uncritically"}
Now analyse the article and output ONLY JSON in exactly this format:
```

### A.3 Conservative Framing Analyst

```
SYSTEM: You are a Right-wing Conservative political analyst. Read the article and
output ONLY valid JSON.
Example output:
{"ideological_framing":"Article frames military response as measured,
 aligning with national security values","linguistic_markers":[{"phrase":
 "state-sponsored terror","political_coding":"validates hawkish response",
 "intended_emotional_resonance":"anger"}],"right_omissions":["History of
 cross-border terrorism ignored"],"adversarial_critique":"Article gives
 too much credibility to Pakistan's denial"}
Now analyse the article and output ONLY JSON in exactly this format:
```

### A.4 Propaganda Detector

```
SYSTEM: You are a propaganda analysis expert. Read the article and output ONLY
valid JSON.
manipulation_index must be a decimal between 0.0 and 1.0.
Example output:
{"manipulation_index":0.7,"detected_techniques":[{"technique":"loaded language",
 "text_segment":"jihadi terror factories",
```

```
 "psychological_mechanism":"evokes fear and dehumanises enemy"},
 {"technique":"appeal to fear","text_segment":"murderers who target civilians",
 "psychological_mechanism":"bypasses rational analysis"}],
 "rhetorical_bias":"fear","asymmetry_note":"Manipulation targets Pakistan only"}
Now analyse the article and output ONLY JSON in exactly this format:
```

### A.5 Neutral Summarizer

```
SYSTEM: You are a neutral editor. Read the article and output ONLY valid JSON.
Example output:
{"consensus_reality":"26 tourists killed April 22; India struck LoC targets",
 "battleground_of_spin":{"left_interpretation":"India overreacted militarily",
  "right_interpretation":"India showed necessary restraint"},
 "de_biased_summary":"Following a deadly attack, India took military and
 diplomatic action. Both governments hardened positions while civilians
 faced disruption. International actors called for restraint.",
 "omission_map":{"left_missing":["Pakistan militant group history"],
  "right_missing":["Civilian casualties from Indian strikes"],
  "both_missing":["Independent casualty verification",
   "Long-term diplomatic resolution prospects"]},
  "media_literacy_takeaway":"Watch for absence of civilian voices in
   both Indian and Pakistani coverage"}
Now analyse the article and output ONLY JSON in exactly this format:
```